# A MINIATURIZED SEMANTIC SEGMENTATION METHOD FOR REMOTE SENSING IMAGE

*Shou-Yu Chen, Guang-Sheng Chen and Wei-Peng Jing*

College of Information and Computer Engineering, Northeast Forestry University, Harbin, China

## ABSTRACT

In order to save the memory, we propose a miniaturization method for neural network to reduce the parameter quantity existed in remote sensing (RS) image semantic segmentation model. The compact convolution optimization method is first used for standard U-Net to reduce the weights quantity. With the purpose of decreasing model performance loss caused by miniaturization and based on the characteristics of remote sensing image, fewer down-samplings and improved cascade atrous convolution are then used to improve the performance of the miniaturized U-Net. Compared with U-Net, our proposed Micro-Net not only achieves 29.26 times model compression, but also basically maintains the performance unchanged on the public dataset. We provide a Keras and Tensorflow hybrid programming implementation for our model: https://github.com/Isnot2bad/Micro-Net

## 1. INTRODUCTION

As the major data source in mapping [1], earth observation [2], ground target recognition [3], RS images have important research value. Different from target recognition, semantic segmentation classifies each image pixel into one semantic category [4, 5] to obtain richer category information. As the RS image resolution gradually increases, larger data amount has put forward new requirement for efficient execution. Since deep learning [6] was proposed in 2006, and thanks to recent advances in computer vision [7, 8], convolutional neural network (CNN) has been extensively used in RS field as an efficient technique compared with traditional machine learning. At present, there are several CNN-based semantic segmentation models, such as FCN [4], U-Net [9] and DeepLab series [10-13]. However, they all have overmuch trainable weights needed to be reduced.

In order to reduce weights quantity, recent studies mainly concentrate on parameter pruning and sharing, low-rank factorization, knowledge distillation, transferred/compact convolutional filters. In the aspect of parameter pruning and sharing, k-means quantization [14] and 8-bit parameter quantization [15] are used for parameter values, and a complete process of deep network compression [16] is proposed. In sparsity and low-rank decomposition respect, tensor decomposition strategies [17] is used to achieve 4.5 times acceleration at the expense of decreasing 1% text recognition accuracy, besides, kernel tensor polynomial decomposition [18] is proposed with nonlinear least squares method. In terms of knowledge distillation, knowledge transfer [19] is used for model compression, another novel compression framework [20] is introduced to reduce the training times of deep network by following "student-teacher" paradigm, in which "students" were punished by softening the output of "teacher". In regard to transferred/compact convolutional filters, Inception [21] decomposed 3x3 convolution into two 1x1 convolutions, SqueezeNet [22] replaced 3x3 convolution with 1x1 convolution and created a compact model with less parameter while maintained network performance compared with AlexNet [23]. Despite various studies for miniaturized classification method, research on miniaturized semantic segmentation model for RS images is still less.

In addition, different from self-driving and indoor scene semantic segmentation data, the target size in remote sensing image is usually smaller, houses for example are scattered, consequently the RS image semantic segmentation model is required to have better segmentation ability for details. Fully convolutional network (FCN) [24] was proposed in 2014, which extended CNN and could be used for semantic segmentation. The succeeding model U-Net [9] originated from FCN, and it is a classic network with clear structure and thus has research value of control experiment. Using pooling layer in U-Net increased receptive field of high-layer neurons on input, however, it lost the exact location information, which is bad for semantic segmentation. The situation is worse when segmenting small target in RS image. In extreme cases, if the target size is less than 16 pixels, the fourth down-sampling's output feature map in U-Net will completely lose information about target, thus increasing the difficulty of semantic segmentation. For this problem, we can consider two solutions: one way is to improve the encoder-decoder model from the macro level, the encoder gradually reduces the spatial dimension of the feature map by down-sampling layers at the expense of losing detailed information, and the decoder progressively restores dimension and detail of the target, moreover, the use of bypass-connection from encoder to decoder usually helps the decoder to restore the object details better. Another solution is improving the network model from the micro level, for instance, replacing pooling layer with atrous convolution [25]. After atrous convolution is proposed, "background module" is put forward which using atrous convolution under multiscale aggregation conditions. Similarly, DeepLab series [10-12] used atrous convolution to enlarge neuron's receptive field without decreasing spatial dimension of feature map, and combined multiscale features to improve segmentation performance. Although the atrous convolution could increase receptive field of high-layer neurons on the input image, however, because of cascade 3x3 atrous convolution with the same rate is widely used in existing networks, receptive field is pore-like distribution actually and do not overlap. For that reason, the discontinuity of semantic information in high-layer feature map causes the point-like noise on the segmentation result. To solve this problem, the cascade approach of atrous convolution need to be adjusted. For similar problems, HDC design structure [26] using different atrous rate is proposed and a similar method [27] is used to explore the solution in the semantic segmentation task further, both of them provide a good idea for improving the cascade atrous convolution.

In this paper, we conduct research on the basis of U-Net because of its classics, clear model structure, simple control experiments, and ease of structural expansion in future study. We propose a novel miniaturized method to reduce weights quantity in semantic segmentation model of RS image. The standard U-Net is miniaturized first. After we analyzed the performance loss introduced by miniaturization, our network structure is optimized step by step. The final model uses fewer down-sampling operations, re-adjusted cascade atrous convolution and deeper encoder to achieves satisfactory experiment results meanwhile its weights quantity is reduced greatly.

## 2. PROPOSED METHOD: MICRO-NET

With the purpose of designing a semantic segmentation artificial neural network with fewer weights, meanwhile reducing the performance loss caused by miniaturization, we use the following two design strategies:

Strategy 1. Model miniaturization: because nearly all trainable weights of U-Net exist in its convolution layers, the compact convolution is used to decrease weights quantity in original convolution layers in U-Net.

Strategy 2. Model performance improvement: for the model performance reduction caused by miniaturization, we analyze the data processing process in U-Net and find out the reason why the detail information is lost, then optimize the model correspondingly.

### 2.1. Model Miniaturization

After analyzed U-Net, we observe that all weights of network exist in the convolution layers, so the weights quantity in these layers needs to be reduced. Because of its good performance as a compact convolution module, we use fire module [22] to miniaturize the network. Unlike SqueezeNet and method in [28], the encoder and decoder in the network are miniaturized with compact convolution because of the purpose of miniaturize semantic segmentation networks, that is, all convolutions in U-Net except the last one are replaced with fire module, which reduces the weights amount greatly.

In fire module, a 1x1 filter in its squeeze convolution layer is first used to reduce the channel number of input feature maps. Then, a 1x1 and a 3x3 filters in expand 1x1 and 3x3 convolution layers are used to convolute on the reduced maps respectively. Finally, the results are merged on the channel dimension and then output. We follow the naming rules in [22] and use $s_{i,1\times1}$, $e_{i,1\times1}$, $e_{i,3\times3}$ to denote the number of filters in squeeze 1x1, expand 1x1 and expand 3x3 layers within the $i$-th fire module respectively.

$e_i = base_e \times 2^{\left\lfloor \frac{i}{freq} \right\rfloor}$ is used to indicate filters amount in expand layer of the $i$-th fire module in encoder, that is, $e_i = e_{i,1\times1} + e_{i,3\times3}$. $base_e$ is the base number of filters in expand layer, and $e_i$ in expand layer is doubled every $freq$ fire modules. We define $p_{3\times3}$. (in range [0, 1], all fire modules share this parameter) as the ratio of $e_{i,3\times3}$ to $e_i$, and $SR$ (squeeze ratio, again, in range (0, 1], shared by all fire modules) as the ratio of filters quantity in squeeze layer to expand layer, that is, $s_{i,1\times1} = SR \times e_i$. In decoder, the order of parameter $e_i$ within each fire module sequence is exactly opposite to their counterpart in encoder. See Section 3.2 for network performance before and after miniaturization.

### 2.2. Performance Optimization

The miniaturized model in Section 2.1 leads to a certain performance loss. We believe that the damage to performance comes mainly from the excessive use of down-sampling in the model. Therefore, we improve the semantic segmentation network from three aspects: reducing the number of down-samplings, using atrous convolution and adjusting the encoder depth.

#### 2.2.1. Reducing the number of down-samplings

First, the last two max-poolings and the first two deconvolutions in U-Net are removed, also the convolutions between these deleted items. Then, the feature map is down-sampled up to 4 times (4x) during the whole calculation process to ensure that the loss of spatial information is reduced. Later, the depth of the model needs to be increased to preserve the network's fitting ability. To achieve that, the number of modules in each fire module sequence (both in miniaturized model's encoder and decoder in Section 2.1) increased from 2 to 3 to sustain the network depth unchanged. Experiment result is shown in Table 2, line 3.

#### 2.2.2. Using atrous convolution

Because the typical cascade atrous convolutions in a group used in mainstream networks usually use the same atrous rate, therefore, for receptive fields on input image of adjacent pixels in high-level hidden layer, there exists no overlapping pixels, which leads easily to noise on the segmentation result and unsatisfactory segmentation effect of small target. To solve this problem, we explore atrous convolution inside each successive fire module within every sequence in encoder and use different values with the greatest common divisor of 1 as atrous convolution rates, that is, the

3x3 convolution in each fire module is replaced by atrous convolution with a specific atrous rate. Similarly, the atrous rates in inverted order are used in atrous convolutions in decoder in accordance with their counterparts in encoder. We attempted different arrangements of atrous rates, the best atrous convolution scheme in encoder and experiment results are shown in Fig.1 and Section 3.2 respectively.

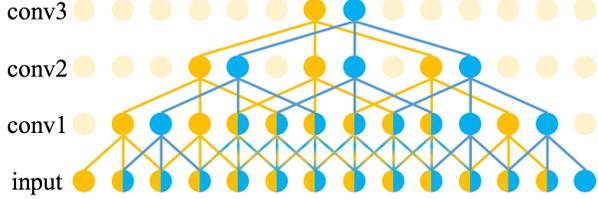

**Fig. 1**. schematic diagram of overlapping receptive field of two adjacent neurons on top-most layer. The atrous rates of above convolutions from top to bottom are 3, 2, 1 respectively.

*2.2.3. Adjusting the encoder depth*

Deeper layers help to improve performance of neural network [23], so we study the depth of decoder. We insert standard fire modules before each improved fire module sequence in the encoder. This modification increases the depth of the encoder and remains the depth of decoder unchanged. The experiment result is shown in Section 3.2.

The structure of final model in this paper is presented in Fig. 2(c), and the model architectural dimensions is given in Table 3. After the input image has been processed by encoder's 3 fire module sequences, which include 3x3 atrous convolution inside, and 2 max-pooling layers, intermediate result is then passed through 3 fire module sequences and 2 deconvolution layers in decoder, its output is then processed by the last convolution layer to generate the final output. At the end of the 1st and 2nd fire module sequences, there have two bypass connections to the start of their counterpart sequences in decoder. Compared with U-Net, our proposed Micro-Net's weights amount is decreased significantly, and the performance is almost unchanged. The model is highly competitive in terms of memory usage.

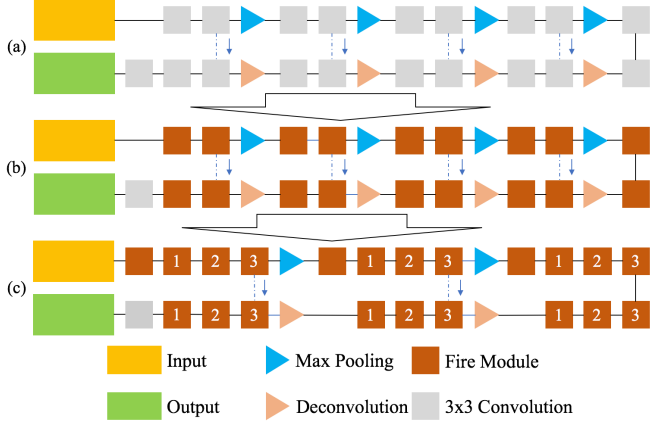

**Fig. 2**. U-Net based structure modification flow chart. (a) U-Net, (b) Miniaturized-Net, (c) Micro-Net. The number inside fire module represent atrous convolution rate determined by experiment. The output of the network is dense prediction map generated on the last convolutional layer using Softmax, which has the same size as the input image.

## 3. EXPERIMENTS AND RESULTS

### 3.1. Dataset & Configuration & Evaluation Indices

Dataset: The dataset used in this paper come from the Austin City image 1-18 in the public dataset [29]. Each image is 5000x5000 pixels in size and has one ground truth image for two semantic categories including building and not building.

Configuration: Stochastic gradient descent (SGD) optimization algorithm is used with batch size equals 2 to train all models in our paper while cross entropy is chosen as loss function:

$$loss = -\sum_i log(y_i log_{a_i}) \quad (1)$$

$y_i$ and $a_i$ represent category value (in one-hot format) and network output of pixel $i$ respectively. Original images are cut into patches with size 500x500, among them, 90% patches are used as training dataset, and the rest as validation. We take $base_e = 64$ according to U-Net [9]. To maintain model's fitting ability, we set $p_{3\times3} = 0.5$. Besides, same as [22], $freq = 2$, $SR = 0.125$. Each model is trained for 20 rounds with an initial learning rate of 0.001, weight decay of 0.00005 and momentum of 0.9. Other details: all convolutions use ReLU as non-linear activation function without bias, and use 'SAME' as convolution type in Tensorflow/Keras which means zero padding equals 1 for all 3x3 convolution. No regularization is used. Only horizontal flip is utilized for data augmentation. The model weights is initialized by the method proposed in [30].

Evaluation Indices: Mean Intersection Over Union (mIOU) and accuracy (ACC) are used to evaluate model performance:

$$mIOU = \frac{1}{n_c}\sum_i \left[n_{ii}/\left(t_i + \sum_j n_{ji} - n_{ii}\right)\right] \quad (2)$$

$$ACC = \sum_i n_{ii} / \sum_i t_i \quad (3)$$

In all images in the validation dataset, $n_c$ is the sum of categories included in ground truth, $n_{ji}$ is the number of pixel category $j$ in label which is predicted as category $i$ in model's output, $t_i$ is pixels quantity of category $i$ in ground truth.

### 3.2. Experiments

All experiment results are obtained on the validation dataset, and compared in terms of weights quantity (Param, in millions), mIOU and ACC.

*3.2.1. Experiment 1. U-Net vs. Miniaturized Net*

The U-Net and miniaturized network in Section 2.1 are used for control experiment to study the effect of miniaturization in terms of network performance. The experiment result is shown in Table 1.

In Table 1, the weights amount in miniaturized network is declined by 5.79 times, which validates the effectiveness of our model miniaturization method, but at the same time, there exists 1% ACC and 3.89% mIOU performance degradation. For convenience of reference below, we rename Miniaturized-Net to Benchmark1, abbreviated as BM1.

**Table 1**. Model performance before and after miniaturization.

| model | Param (M) | mIOU (%) | ACC (%) | Alias |
|---|---|---|---|---|
| U-Net | 31.02 | 79.48 | 95.54 | |
| Miniaturized-Net | 5.36 | 75.59 | 94.54 | BM1 |

*3.2.2. Experiment 2. Miniaturized-Net vs. Micro-Net*

we first reduce the number of down-samplings to 2. Experiment result is presented in Table 2, line 3. The result shows that reducing the number of down-samplings has a certain improvement on network performance. We suppose that the result of mIOU rising and ACC falling come mainly from the unbalanced category distribution of dataset. Furthermore, since there are less filters in the newly added fire modules than the deleted ones, the weights quantity in the network is reduced further. We rename this model as Benchmark2, abbreviated as BM2.

**Table 2**. Network performance optimization experiment results. '4x U-Net' represents the model with 2 max-poolings in Section 2.2.1. '(1,2,3), (1,2,3), (1,2,3)' in line 4 indicates that the 1st, 2nd, 3rd fire module sequences in encoder using atrous convolution sequence with atrous convolution rate of 1, 2, 3. '(1,1,2,3)' in line 6 indicates the number of modules in each fire module sequence in encoder is increased to 4, and the atrous convolution rate used by modules within each sequence is 1,1,2,3 respectively. The number of modules in each sequence in decoder remains unchanged.

| model | Param (M) | mIOU (%) | ACC (%) | Alias |
|---|---|---|---|---|
| Miniaturized Net | 5.36 | 75.59 | 94.54 | BM1 |
| 4x U-Net | 0.93 | 75.65 | 94.42 | BM2 |
| (1,2,3), (1,2,3), (1,2,3) | 0.93 | **75.78** | 94.40 | BM3 |
| (1,2,5), (1,2,3), (1,1,2) | 0.93 | 75.70 | 94.52 | |
| (1,1,2,3) | 1.06 | **76.81** | 94.79 | Micro-Net |
| (1,1,1,2,3) | 1.18 | 76.37 | 94.69 | |

We use the improved cascade atrous convolution strategy proposed in Section 2.2 to explore the different atrous convolution rates meanwhile keep other configures unchanged. The results in Table 2, line 4 and 5 verify the effectiveness of our strategy. We rename the model in Table 2, which achieve the best mIOU using atrous rate (1,2,3), (1,2,3), (1,2,3), to Benchmark 3, abbreviated as BM3.

Then, we explore the impact of encoder depth on model performance. We try different encoder depths, the experiment results are shown in Table 2, line 6 and 7, and indicate that increasing depth of the encoder helps to improve model's semantic segmentation performance. The model with atrous convolution rate (1,1,2,3) in encoder achieved the best mIOU and is finally renamed as Micro-Net. We believe that the reason why mIOU in row 7 is lower than in row 6 is that the model over-fitted on the training dateset. Besides, due to the newly added network layer, the weights quantity of Micro-Net is increased slightly compared with BM3.

Compared with U-Net, Micro-Net's ACC and mIOU decreased acceptably by 0.75% and 2.67% respectively, within the acceptable range, yet weights quantity decreased by 29.26 times noticeably. Experiment results prove validity of our method. The performance comparison chart of several key networks in our paper is shown in Fig. 3, and architectural dimensions of Micro-Net is listed in Table 3:

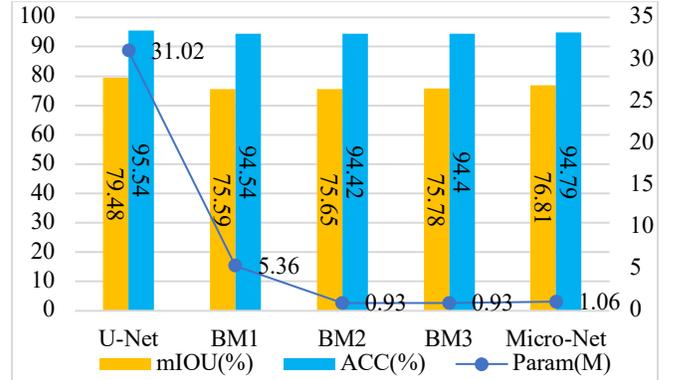

**Fig. 3.** Performance comparison of key models.

| layer name | output feature map size | depth | s1x1 | e1x1 | e3x3 | Param |
|---|---|---|---|---|---|---|
| input | 500x500x3 | | | | | |
| fm 1 | 500x500x64 | 2 | 16 | 32 | 32 | 5158 |
| fm 2~4 | 500x500x64 | 2 | 16 | 32 | 32 | 6144 |
| mp 1 | 250x250x64 | 0 | | | | |
| fm 5 | 250x250x128 | 2 | 32 | 64 | 64 | 22528 |
| fm 6~8 | 250x250x128 | 2 | 32 | 64 | 64 | 24576 |
| mp 2 | 125x125x128 | 0 | | | | |
| fm 9 | 125x125x256 | 2 | 64 | 128 | 128 | 90112 |
| fm 10~12 | 125x125x256 | 2 | 64 | 128 | 128 | 98304 |
| dfm 9~7 | 125x125x256 | 2 | 64 | 128 | 128 | 98304 |
| dec 1 | 250x250x128 | 1 | | | | 131072 |
| add 1 | 250x250x128 | | | | | |
| dfm 6~4 | 250x250x128 | 2 | 32 | 64 | 64 | 24576 |
| dec 2 | 500x500x64 | 1 | | | | 32768 |
| add 2 | 500x500x64 | | | | | |
| dfm 3~1 | 500x500x64 | 2 | 16 | 32 | 32 | 6144 |
| conv | 500x500x2 | 2 | | | | 128 |

**Table. 3**. Architectural dimensions of Micro-Net. 'fm', 'mp', 'dfm', 'dec', 'conv' represents 'fire module', 'max pooling', 'fire module in decoder', 'deconvolution', 'convolution' respectively. The choice of parameters s1x1, e1x1 and e3x3 is explained in Section 2.1 and 3.1.

## 4. CONCLUSIONS

In order to lower the memory usage of semantic segmentation network, we adopt some model optimization methods. At the micro level, we use compact convolution and improved cascade atrous convolution. At the macro level, we reduce the number of down-samplings and increase the depth of encoder. Eventually, the proposed Micro-Net achieved 94.79% ACC and 76.81% mIOU on the public dataset, meanwhile its model size dropped by 29.26 times.

# 5. REFERENCES

[1] Zhao, W., S. Du, and W.J. Emery, "Object-Based Convolutional Neural Network for High-Resolution Imagery Classification". IEEE Journal of Selected Topics in Applied Earth Observations & Remote Sensing, vol. PP-99, p. 1-11, 2017.

[2] Volpi, M. and M. Kanevski, "Supervised Change Detection in VHR Images Using Support Vector Machines and Contextual Information". International Journal of Applied Earth Observation & Geoinformation, vol. 20-2, p. 77–85, 2010.

[3] Guan, H., et al., "Deep learning-based tree classification using mobile LiDAR data". Remote Sensing Letters, vol. 6-11, p. 864-873, 2015.

[4] Long, J., E. Shelhamer, and T. Darrell, "Fully convolutional networks for semantic segmentation". IEEE Transactions on Pattern Analysis & Machine Intelligence, vol. 39-4, p. 640, 2017.

[5] Marmanis, D., et al., "Semantic Segmentation of Aerial Images with AN Ensemble of Cnns". vol. III-3, p. 473-480, 2016.

[6] Hinton, G.E. and R.R. Salakhutdinov, "Reducing the dimensionality of data with neural networks". Science, vol. 313-5786, p. 504-7, 2006.

[7] Krizhevsky, A., I. Sutskever, and G.E. Hinton. "Imagenet classification with deep convolutional neural networks". of Conference. Year.

[8] Simonyan, K. and A. Zisserman, "Very Deep Convolutional Networks for Large-Scale Image Recognition". CoRR, vol. abs/1409.1556, 2014.

[9] Ronneberger, O., P. Fischer, and T. Brox. "U-Net: Convolutional Networks for Biomedical Image Segmentation". in Medical Image Computing and Computer-Assisted Intervention – MICCAI 2015. of Conference. Cham: Springer International Publishing. Year.

[10] Chen, L.-C., et al. "Semantic Image Segmentation with Deep Convolutional Nets and Fully Connected CRFss". ArXiv e-prints, 1412. 2014.

[11] Chen, L.C., et al., "DeepLab: Semantic Image Segmentation with Deep Convolutional Nets, Atrous Convolution, and Fully Connected CRFs". IEEE Transactions on Pattern Analysis and Machine Intelligence, vol. 40-4, p. 834-848, 2018.

[12] Chen, L.-C., et al. "Rethinking Atrous Convolution for Semantic Image Segmentations". ArXiv e-prints, 1706. 2017.

[13] Chen, L.-C., et al. "Encoder-Decoder with Atrous Separable Convolution for Semantic Image Segmentations". ArXiv e-prints, 1802. 2018.

[14] Gong, Y., et al. "Compressing Deep Convolutional Networks using Vector Quantizations". ArXiv e-prints, 1412. 2014.

[15] Vanhoucke, V. and M.Z. Mao, "Improving the speed of neural networks on CPUs". Deep Learning & Unsupervised Feature Learning Workshop Nips, vol., 2011.

[16] Han, S., H. Mao, and W.J. Dally "Deep Compression: Compressing Deep Neural Networks with Pruning, Trained Quantization and Huffman Codings". ArXiv e-prints, 1510. 2015.

[17] Sironi, A., et al. "Learning Separable Filters". in Computer Vision and Pattern Recognition. of Conference. Year.

[18] Lebedev, V., et al. "Speeding-up Convolutional Neural Networks Using Fine-tuned CP-Decompositions". ArXiv e-prints, 1412. 2014.

[19] Caruana, R. and A. Niculescu-Mizil. "Model compression". in ACM SIGKDD International Conference on Knowledge Discovery and Data Mining. of Conference. Year.

[20] Hinton, G., O. Vinyals, and J. Dean "Distilling the Knowledge in a Neural Networks". ArXiv e-prints, 1503. 2015.

[21] Szegedy, C., et al., "Inception-v4, Inception-ResNet and the Impact of Residual Connections on Learning". vol., 2016.

[22] Iandola, F.N., et al. "SqueezeNet: AlexNet-level accuracy with 50x fewer parameters and <0.5MB model sizes". ArXiv e-prints, 1602. 2016.

[23] Krizhevsky, A., I. Sutskever, and G.E. Hinton. "ImageNet classification with deep convolutional neural networks". in International Conference on Neural Information Processing Systems. of Conference. Year.

[24] Kang, K. and X. Wang, "Fully convolutional neural networks for crowd segmentation". arXiv preprint arXiv:1411.4464, vol., 2014.

[25] Yu, F. and V. Koltun "Multi-Scale Context Aggregation by Dilated Convolutionss". ArXiv e-prints, 1511. 2015.

[26] Wang, P., et al. "Understanding Convolution for Semantic Segmentations". ArXiv e-prints, 1702. 2017.

[27] Hamaguchi, R., et al. "Effective Use of Dilated Convolutions for Segmenting Small Object Instances in Remote Sensing Imagerys". ArXiv e-prints, 1709. 2017.

[28] Treml, M., et al. "Speeding up Semantic Segmentation for Autonomous Driving". in NIPS 2016 Workshop - MLITS. of Conference. Year.

[29] Maggiori, E., et al. "Can semantic labeling methods generalize to any city? the inria aerial image labeling benchmark". in IGARSS 2017 - 2017 IEEE International Geoscience and Remote Sensing Symposium. of Conference. Year.

[30] He, K., et al., "Delving Deep into Rectifiers: Surpassing Human-Level Performance on ImageNet Classification". vol., p. 1026-1034, 2015.